\documentclass[letterpaper]{article} 
\usepackage[submission]{aaai23}  
\usepackage{times}  
\usepackage{helvet}  
\usepackage{courier}  
\usepackage[hyphens]{url}  
\usepackage{graphicx} 
\usepackage{natbib}  
\usepackage{caption} 
\usepackage{algorithm}
\usepackage{algorithmic}

\usepackage{microtype}
\usepackage{graphicx}
\usepackage{caption}
\usepackage{subcaption}
\usepackage{booktabs} 
\usepackage{tikz-cd}
\usetikzlibrary{arrows}
\usepackage{amsmath}
\usepackage{comment}
\usepackage{amsthm}
\usepackage{amssymb}
\usepackage{url}
\usepackage{tikz}
\usepackage{tikz-qtree,tikz-qtree-compat}
\usepackage{soul}

\makeatletter
\newtheorem*{rep@theorem}{\rep@title}
\newcommand{\newreptheorem}[2]{%
\newenvironment{rep#1}[1]{%
 \def\rep@title{#2 \ref{##1}}%
 \begin{rep@theorem}}%
 {\end{rep@theorem}}}
\makeatother

\newtheorem{theorem}{Theorem}
\newreptheorem{theorem}{Theorem}

\usepackage{stackengine}


\usepackage{newfloat}
\usepackage{listings}
\DeclareCaptionStyle{ruled}{labelfont=normalfont,labelsep=colon,strut=off} 
\floatstyle{ruled}
\newfloat{listing}{tb}{lst}{}
\floatname{listing}{Listing}
\title{Unit Selection with Nonbinary Treatment and Effect}
\author{
    Ang Li,
    Judea Pearl
}
\affiliations{
    Cognitive Systems Laboratory, Department of Computer Science,\\
    University of California, Los Angeles,\\
    Los Angeles, California, USA.\\
    \{angli, judea\}@cs.ucla.edu
}

\usepackage{bibentry}

\begin{document}

\maketitle

\begin{abstract}
The unit selection problem aims to identify a set of individuals who are most likely to exhibit a desired mode of behavior, for example, selecting individuals who would respond one way if encouraged and a different way if not encouraged. Using a combination of experimental and observational data, Li and Pearl derived tight bounds on the “benefit function”, which is the payoff/cost associated with selecting an individual with given characteristics. This paper extends the benefit function to the general form such that the treatment and effect are not restricted to binary. We propose an algorithm to test the identifiability of the nonbinary benefit function and an algorithm to compute the bounds of the nonbinary benefit function using experimental and observational data.
\end{abstract}

\section{Introduction}
Several areas of industry, marketing, and health science face the unit selection dilemma. For example, in customer relationship management \cite{berson1999building, lejeune2001measuring, hung2006applying, tsai2009customer}, it is useful to determine the customers who are going to leave but might reconsider if encouraged to stay. Due to the high expense of such initiatives, management is forced to limit inducement to customers who are most likely to exhibit the behavior of interest. As another example, companies are interested in identifying users who would click on an advertisement if and only if it is highlighted in online advertising \cite{yan2009much, bottou2013counterfactual, li2014counterfactual, sun2015causal}. The challenge in identifying these users stems from the fact that the desired response pattern is not observed directly but rather is defined counterfactually in terms of what the individual would do under hypothetical unrealized conditions. For example, when we observe that a user has clicked on a highlighted advertisement, we do not know whether they would click on that same advertisement if it were not highlighted. 

The binary benefit function for the unit selection problem was defined by Li and Pearl \cite{li:pea19-r488} (we will call this Li-Pearl's model), and it properly captures the nature of the desired behavior. Using a combination of experimental and observational data, Li and Pearl derived tight bounds of the benefit function. The only assumption is that the treatment has no effect on the population-specific characteristics. Inspired by the idea of Mueller, Li, and Pearl \cite{pearl:etal21-r505} and Dawid et al. \cite{dawid2017} that the bounds of probabilities of causation could be narrowed using covariates information, Li and Pearl \cite{li2022unit} narrowed the bounds of the benefit function using covariates information and their causal structure. However, the abovementioned studies are based on binary treatment and effect. Recently, researchers have shown interest in developing bounds for probabilities of causation with nonbinary treatment and effect. Zhang, Tian, and Bareinboim \cite{zhang2022partial}, as well as Li and Pearl \cite{li2022bounds}, proposed nonlinear programming-based solutions to compute the bounds of nonbinary probabilities of causation numerically. Li and Pearl \cite{li:pea-r516} provided the theoretical bounds of nonbinary probabilities of causation. The benefit function is a linear combination of probabilities of causation; therefore, in this paper, we focus on discovering the bounds of any benefit function without restricting them to binary treatment and effect.

Consider the following motivating scenario: a clinical study is conducted to test the effectiveness of a vaccine. The treatments include vaccinated and unvaccinated. The outcomes include uninfected, asymptomatic infected, and infected in a severe condition. The benefited individuals include the following: the individual who would be infected in a severe condition if unvaccinated and would be asymptomatic infected if vaccinated, the individual who would be infected in a severe condition if unvaccinated and would be uninfected if vaccinated, and the individual who would be asymptomatic infected if unvaccinated and would be uninfected if vaccinated. The harmed individuals include the following: the individual who would be asymptomatic infected if unvaccinated and would be infected in a severe condition if vaccinated, the individual who would be uninfected if unvaccinated and would be infected in a severe condition if vaccinated, and the individual who would be uninfected if unvaccinated and would be asymptomatic infected if vaccinated. All others are unaffected individuals. The researcher performing the clinical study has collected both experimental and observational data. The researcher then wants to know the expected difference between benefited and harmed individuals to emphasize the effectiveness of the vaccine.

We cannot apply Li-Pearl's model because we have two treatments and three outcomes. In this paper, we extend Li-Pearl's benefit function to general form without restricting them to binary treatment and effect. We will provide an algorithm to test the identifiability of the nonbinary benefit function and an algorithm to compute the bounds of the nonbinary benefit function using experimental and observational data.

\section{Preliminaries}
\label{related work}
In this section, we review Li and Pearl's binary benefit function of the unit selection problem \cite{li:pea19-r488}, and the theoretical bounds of the probabilities of causation recently proposed by Li and Pearl \cite{li:pea-r516}. 

In this paper we use the language of counterfactuals in structural model semantics, as given in \cite{galles1998axiomatic,halpern2000axiomatizing}. we use $Y_x=y$ to denote the counterfactual sentence ``Variable $Y$ would have the value $y$, had $X$ been $x$". For simplicity purposes, in the rest of the paper, we use $y_x$ to denote the event $Y_x=y$, $y_{x'}$ to denote the event $Y_{x'}=y$, $y'_x$ to denote the event $Y_x=y'$, and $y'_{x'}$ to denote the event $Y_{x'}=y'$. we assume that experimental data will be summarized in the form of the causal efects such as $P(y_x)$ and observational data will be summarized in the form of the joint probability function such as $P(x,y)$. If not specified, the variable $X$ stands for treatment and the variable $Y$ stands for effect.

Individual behavior was classified into four response types: labeled complier, always-taker, never-taker, and defier. Suppose the benefit of selecting one individual in each category are $\beta, \gamma, \theta, \delta$ respectively (i.e., the benefit vector is $(\beta, \gamma, \theta, \delta)$). Li and Pearl defined the objective function of the unit selection problem as the average benefit gained per individual. Suppose $x$ and $x'$ are binary treatments, $y$ and $y'$ are binary outcomes, and $c$ are population-specific characteristics, the objective function (i.e., benefit function) is following (If the goal is to evaluate the average benefit gained per individual for a specific population $c$, $argmax_c$ can be dropped.):
\begin{eqnarray*}
\label{liobj}
argmax_c \text{ }\beta P(y_{x},y'_{x'}|c)+\gamma P(y_{x},y_{x'}|c) + \nonumber \\+\theta P(y'_{x},y'_{x'}|c)+\delta P(y'_{x},y_{x'}|c).
\end{eqnarray*}
Using a combination of experimental and observational data, Li and Pearl established the most general tight bounds on this benefit function (which we refer to as Li-Pearl's Theorem in the rest of the paper). The only constraint is that the population-specific characteristics are not a descendant of the treatment.

Li and Pearl \cite{li:pea-r516} provided eight theorems to compute bounds for any type of probabilities of causation with nonbinary treatment and effect. Suppose variable $X$ has $m$ values and $Y$ has $n$ values, the following probabilities of causation are bounded. Besides, if the probabilities of causation are conditioned on a population-specific variable $c$ that is not affected by $X$, then all the theorems still hold (we provided the extended theorems from Li and Pearl in the appendix).

\begin{eqnarray*}
&&P({y_i}_{x_j}, y_i), \\
&&s.t., 1 \le i \le n, 1 \le j \le m,\\
&&P({y_i}_{x_j}, y_k), \\
&&s.t., 1 \le i,k \le n, 1 \le j \le m, i\ne k\\
&&P({y_i}_{x_j}, x_k), \\
&&s.t., 1 \le i \le n, 1 \le j,k \le m, j\ne k\\
&&P({y_i}_{x_j}, y_k, x_p), \\
&&s.t., 1 \le i,k \le n, 1 \le j,p \le m, j\ne p\\
&&P({y_{i_1}}_{x_{j_1}},...,{y_{i_k}}_{x_{j_k}}), \\
&&s.t., 1 \le i_1,...,i_k \le n, 1 \le j_1,...,j_k \le m, j_1\ne ... \ne j_k\\
&&P({y_{i_1}}_{x_{j_1}},...,{y_{i_k}}_{x_{j_k}},x_p), \\
&&s.t., 1 \le i_1,...,i_k \le n, 1 \le j_1,...,j_k,p \le m,\\
&&j_1\ne ... \ne j_k \ne p\\
&&P({y_{i_1}}_{x_{j_1}},...,{y_{i_k}}_{x_{j_k}},y_q), \\
&&s.t., 1 \le i_1,...,i_k,q \le n, 1 \le j_1,...,j_k \le m,\\
&&j_1\ne ... \ne j_k\\
&&P({y_{i_1}}_{x_{j_1}},...,{y_{i_k}}_{x_{j_k}},x_p,y_q),\\
&&s.t., 1 \le i_1,...,i_k,q \le n, 1 \le j_1,...,j_k,p \le m,\\
&&j_1\ne ... \ne j_k \ne p.
\end{eqnarray*}

The benefit function is a linear combination of the probabilities of causation; therefore, we define the general benefit function for the unit selection problem based on Li and Pearl's results.

\section{Counterfactual Formulation of the Unit Selection Problem}
Based on Li and Pearl \cite{li:pea19-r488}, the objective is to find a set of characteristics $c$ that maximizes the benefit associated with the resulting mixture of different response types of individuals. Let $X$ denotes the treatment with $m$ values and $Y$ denotes the effect with $n$ values. Therefore, we have $n^m$ different response types (i.e., one response type means assigning one effect to each of the treatments). Suppose the benefit of selecting an individual are $(\alpha_1,...,\alpha_{n^m})$ (we call $(\alpha_1,...,\alpha_{n^m})$ as benefit vector). Our objective, then, should be to find $c$ that maximizes the following expression (If the goal is to evaluate the average benefit gained per individual for a specific population $c$, $argmax_c$ can be dropped):

\begin{eqnarray*}
argmax_c && \alpha_1 P({y_1}_{x_1},{y_1}_{x_2},...,{y_1}_{x_m}|c)+\nonumber\\
&&\alpha_2 P({y_1}_{x_1},{y_1}_{x_2},...,{y_2}_{x_m}|c)+... \nonumber \\
&&\alpha_n P({y_1}_{x_1},{y_1}_{x_2},...,{y_n}_{x_m}|c)+... \nonumber \\
&&\alpha_{n^{m-1}+1} P({y_2}_{x_1},{y_1}_{x_2},...,{y_1}_{x_m}|c)+...\nonumber\\
&&\alpha_{n^m} P({y_n}_{x_1},{y_n}_{x_2},...,{y_n}_{x_m}|c).
\label{eqobj}
\end{eqnarray*}

Note that $c$ can be interpreted as the population-specific variable, the only assumption is that the treatment $X$ has no effect on the population-specific variable. Recall from Li and Pearl's paper \cite{li:pea19-r488}, the benefit vector is provided by the decision-maker who uses the model.

In the next section, we will provide an algorithm that could check whether a given benefit function with the benefit vector is identifiable with purely experimental data (i.e., we can find the exact value of the benefit function rather than bounds). If it is not identifiable we will then provide an algorithm that computes the bounds of the benefit function given the benefit vector using experimental and observational data.   

\section{Main Results}
\subsection{Identifiability of Benefit Function}
Recall that in binary case, the conditions of identifiable are gain equality (i.e., $\beta+\delta=\gamma+\theta$) or monotonicity (i.e., $P(y_{x'}, y'_{x})=0$) \cite{li:pea19-r488}. Here, it is complicated in nonbinary cases, therefore; we provide an algorithm to test whether a given benefit function with the benefit vector is identifiable with purely experimental data.

\begin{theorem}
\label{thm1}
Suppose variables $X$ has $m$ values $x_1,...,x_m$ and $Y$ has $n$ values $y_1,...,y_n$. Then the benefit function $f(c)$ is identifiable if Algorithm \ref{alg1} returns (True, res), and res is the value of the benefit function.\\
\begin{eqnarray*}
f(c) =&& \alpha_1 P({y_1}_{x_1},{y_1}_{x_2},...,{y_1}_{x_m}|c)+\nonumber\\
&&\alpha_2 P({y_1}_{x_1},{y_1}_{x_2},...,{y_2}_{x_m}|c)+... \nonumber \\
&&\alpha_n P({y_1}_{x_1},{y_1}_{x_2},...,{y_n}_{x_m}|c)+... \nonumber \\
&&\alpha_{n^{m-1}+1} P({y_2}_{x_1},{y_1}_{x_2},...,{y_1}_{x_m}|c)+...\nonumber\\
&&\alpha_{n^m} P({y_n}_{x_1},{y_n}_{x_2},...,{y_n}_{x_m}|c).
\end{eqnarray*}

\begin{algorithm}
\caption{Check identifiability of the benefit function}
\label{alg1}
\textbf{Input}: $a$, the benefit function,where $a[i]$ is a $m+1$ tuple that stands for ith term in the benefit function. If the ith term is $\alpha_i P({y_{i_1}}_{x_1},{y_{i_2}}_{x_2},...,{y_{{i_m}}}_{x_m}|c)$, then $a[i]=(\alpha_i,i_1,i_2,...,i_m).$\\

$d[1,...,m][1,...,n]$, the experimental data, where $d[i][j]=P({y_j}_{x_i}|c)$.\\

$e$, the adjusted value of the benefit function.\\

The initial call of the algorithm is $IBF(a[1,...,n^m], d[1,...,m][1,...,n], 0)$, where $a[1,...,n^m]$ corresponding to the original benefit function.\\

All lists in this algorithm start with index $1$.\\

\textbf{Output}: (identifiable, value), a tuple, where identifiable $=$ True if the given benefit function is identifiable and value is the value of the benefit function.\\
\\
Function $IBF(a,d,e)$:
\begin{algorithmic}[1] 
\STATE $m=True$;
\STATE $l=length(a)$;
\STATE // Base case, if all benefit vector equals to $(0,...,0)$, then the input benefit function is identifiable, and its value equals to the adjusted value.
\FOR {$i=1$ to $l$}
    \IF {$a[i][1] \ne 0$}
    \STATE $m=False$;
    \STATE break;
    \ENDIF
\ENDFOR
\IF {$m == True$}
    \STATE Return$(True, e)$;
\ENDIF
\STATE // Build an equivalent benefit function by the fact that if $\exists 2\le r\le (m+1) s.t., a[j_1][r]=...=a[j_{n^{m-1}}][r]$, then the sum of these $n^{m-1}$ terms without coefficients is equal to $P({y_{a[j_1][r]}}_{x_{r}})$, we then recursively solve the equivalent benefit function.
\FOR {every $n^{m-1}$ pair in $a$, $(a[j_1],...,a[j_{n^{m-1}}])$, s.t., there $\exists 2\le r\le (m+1) s.t., a[j_1][r]=...=a[j_{n^{m-1}}][r]$}
    \FOR {$k=1$ to $n^{m-1}$}
        \STATE $na =a$;
        \STATE $nc =e + a[j_k][1] * d[r-1][a[j_1][r]]$;
        \FOR {$t=1$ to $n^{m-1}$}
            \IF {$t \ne k$}
                \STATE $na[j_t][1] =na[j_t][1]-na[j_k][1]$;
            \ENDIF
        \ENDFOR
        \STATE $Remove(na[j_k])$;
        \STATE $res = IBF(na, d, nc)$;
        \IF {$res[0] == True$}
            \STATE Return $res$
        \ENDIF
    \ENDFOR
\ENDFOR
\STATE Return $(False, e)$;
\end{algorithmic}
\end{algorithm}
\end{theorem}

The correctness of the algorithm simply follow the fact that $\sum_{n^{m-1}\text{terms}} P(...,{y_i}_{x_j},...|c) = P({y_i}_{x_j}|c)$. Therefore, if there exist such $n^{m-1}$ terms in the benefit function, then we can obtain an equivalent benefit function by replacing one of the $n^{m-1}$ terms with experimental data $P({y_i}_{x_j}|c)$. We exhausted all equivalent benefit functions to check if we could replace all the counterfactual terms with experimental data (i.e., identifiable).

For example, consider $m=n=2$ and the benefit function:
\begin{eqnarray*}
&&7P({y_1}_{x_1},{y_1}_{x_2}|c) + 2P({y_1}_{x_1},{y_2}_{x_2}|c) +\\ &&4P({y_2}_{x_1},{y_1}_{x_2}|c) - P({y_2}_{x_1},{y_2}_{x_2}|c) \\
&=&7P({y_1}_{x_1}|c)-5P({y_1}_{x_1},{y_2}_{x_2}|c)+\\
&&4P({y_2}_{x_1},{y_1}_{x_2}|c) - P({y_2}_{x_1},{y_2}_{x_2}|c) \\
&=&7P({y_1}_{x_1}|c)-5P({y_1}_{x_1},{y_2}_{x_2}|c)+\\
&&4P({y_2}_{x_1}|c) - 5P({y_2}_{x_1},{y_2}_{x_2}|c) \\
&=&7P({y_1}_{x_1}|c)-5P({y_2}_{x_2})+4P({y_2}_{x_1}|c).\\
\end{eqnarray*}
\subsection{Bounds of Benefit Function}
If Algorithm \ref{alg1} returns false, we then need to compute the bounds of the benefit function using experimental and observational data. We first obtain the bounds of the probabilities of causation, $P({y_1}_{x_1},{y_1}_{x_2},...,{y_1}_{x_m}|c)$, ..., $P({y_1}_{x_1},{y_1}_{x_2},...,{y_n}_{x_m}|c)$, ..., $P({y_2}_{x_1},{y_1}_{x_2},...,{y_1}_{x_m}|c)$, ..., $P({y_n}_{x_1},{y_n}_{x_2},...,{y_n}_{x_m}|c)$, by Li and Pearl's theorems \cite{li:pea-r516}. We then have the following theorem. 

\begin{theorem}
\label{thm2}
Suppose variables $X$ has $m$ values $x_1,...,x_m$ and $Y$ has $n$ values $y_1,...,y_n$. Then the bounds of the benefit function $f(c)$ is obtained by Algorithm \ref{alg2}.\\
\begin{eqnarray*}
f(c) =&& \alpha_1 P({y_1}_{x_1},{y_1}_{x_2},...,{y_1}_{x_m}|c)+\nonumber\\
&&\alpha_2 P({y_1}_{x_1},{y_1}_{x_2},...,{y_2}_{x_m}|c)+... \nonumber \\
&&\alpha_n P({y_1}_{x_1},{y_1}_{x_2},...,{y_n}_{x_m}|c)+... \nonumber \\
&&\alpha_{n^{m-1}+1} P({y_2}_{x_1},{y_1}_{x_2},...,{y_1}_{x_m}|c)+...\nonumber\\
&&\alpha_{n^m} P({y_n}_{x_1},{y_n}_{x_2},...,{y_n}_{x_m}|c).
\end{eqnarray*}

\begin{algorithm}
\caption{Compute the bounds of the benefit function}
\label{alg2}
\textbf{Input}: $a$, the benefit function,where $a[i]$ is a $m+1$ tuple that stands for ith term in the benefit function. If the ith term is $\alpha_i P({y_{i_1}}_{x_1},{y_{i_2}}_{x_2},...,{y_{{i_m}}}_{x_m}|c)$, then $a[i]=(\alpha_i,i_1,i_2,...,i_m).$\\

$lb$, the lower bound of all possible terms obtained from Li-Pearl's theorems, where $lb[(i_1,i_2,...,i_m)]$ is the lower bound of $P({y_{i_1}}_{x_1},{y_{i_2}}_{x_2},...,{y_{{i_m}}}_{x_m}|c)$.\\

$ub$, the upper bound of all possible terms obtained from Li-Pearl's theorems, where $ub[(i_1,i_2,...,i_m)]$ is the upper bound of $P({y_{i_1}}_{x_1},{y_{i_2}}_{x_2},...,{y_{{i_m}}}_{x_m}|c)$.\\

$e$, the adjusted value of the benefit function.\\
The initial call of the algorithm is $BBF(a[1,...,n^m], lb, ub, 0)$, where $a[1,...,n^m]$ corresponding to the original benefit function.\\

All lists in this algorithm start with index $1$.\\

\textbf{Output}: (lo, up), lower and upper bound of the benefit function.\\
\\
Function $BBF(a,lb,ub,e)$:
\begin{algorithmic}[1] 
\STATE $l=length(a)$;
\STATE // Base case, compute the bounds.
\STATE $up=e, lo=e$;
\FOR {$i=1$ to $l$}
    \IF {$a[i][1] < 0$}
        \STATE $lo = lo + a[i][1]*ub[(a[i][2],...,a[i][m+1])]$;
        \STATE $up = up + a[i][1]*lb[(a[i][2],...,a[i][m+1])]$;
    \ELSE
        \STATE $lo = lo + a[i][1]*lb[(a[i][2],...,a[i][m+1])]$;
        \STATE $up = up + a[i][1]*ub[(a[i][2],...,a[i][m+1])]$;
    \ENDIF
\ENDFOR
\STATE // Build an equivalent benefit function by the fact that if $\exists 2\le r\le (m+1) s.t., a[j_1][r]=...=a[j_{n^{m-1}}][r]$, then the sum of these $n^{m-1}$ terms without coefficients is equal to $P({y_{a[j_1][r]}}_{x_{r}})$, we then recursively solve the equivalent benefit function.
\FOR {every $n^{m-1}$ pair in $a$, $(a[j_1],...,a[j_{n^{m-1}}])$, s.t., there $\exists 2\le r\le (m+1) s.t., a[j_1][r]=...=a[j_{n^{m-1}}][r]$}
    \FOR {$k=1$ to $n^{m-1}$}
        \STATE $na =a$;
        \STATE $nc =e + a[j_k][1] * d[r-1][a[j_1][r]]$;
        \FOR {$t=1$ to $n^{m-1}$}
            \IF {$t \ne k$}
                \STATE $na[j_t][1] =na[j_t][1]-na[j_k][1]$;
            \ENDIF
        \ENDFOR
        \STATE $Remove(na[j_k])$;
        \STATE $res = BBF(na, lb, ub, nc)$;
        \STATE $lo = \max\{lo, res[0]\}$
        \STATE $up = \min\{up, res[1]\}$
    \ENDFOR
\ENDFOR
\STATE Return $(lo, up)$;
\end{algorithmic}
\end{algorithm}
\end{theorem}

Again, the correctness of the algorithm simply follow the fact that $\sum_{n^{m-1}\text{terms}} P(...,{y_i}_{x_j},...|c) = P({y_i}_{x_j}|c)$. We exhausted all equivalent benefit functions and take the maximum of all the lower bounds and take the minimum of all the upper bounds of equivalent benefit functions.

\section{Example: Effectiveness of a Vaccine}
Recall the motivating example at the beginning, a clinical study is conducted to test the effectiveness of a vaccine. The treatments include vaccinated and unvaccinated. The outcomes include uninfected, asymptomatic infected, and infected in a severe condition. The researcher of the clinical study has collected both experimental and observational data. 

\subsection{Task 1}
The researcher wants to know the expected difference between benefited and harmed individuals to emphasize the effectiveness of the vaccine.

Let $X$ denotes vaccination with $x_1$ being vaccinated and $x_2$ being unvaccinated and $Y$ denotes outcome, where $y_1$ denotes uninfected, $y_2$ denotes asymptomatic infected, and $y_3$ denotes infected in a severe condition. The experimental and observational data of the clinical study are summarized in Tables \ref{tb1} and \ref{tb2}.

\begin{table}
\centering
\begin{tabular}{|c|c|c|}
\hline 
&Vaccinated&Unvaccinated\\
\hline
Uninfected&\begin{tabular}{c}$52$\\People\end{tabular}&\begin{tabular}{c}$329$\\People\end{tabular}\\
\hline
Asymptomatic&\begin{tabular}{c}$512$\\People\end{tabular}&\begin{tabular}{c}$58$\\People\end{tabular}\\
\hline
Severe Condition&\begin{tabular}{c}$36$\\People\end{tabular}&\begin{tabular}{c}$213$\\People\end{tabular}\\
\hline
Overall&\begin{tabular}{c}$600$\\People\end{tabular}&\begin{tabular}{c}$600$\\People\end{tabular}\\
\hline
\end{tabular}
\caption{Experimental data of the clinical study. Here, $600$ people were forced to take the vaccine and $600$ people were forced to take no vaccine.}
\label{tb1}
\end{table}

\begin{table}
\centering
\begin{tabular}{|c|c|c|}
\hline 
&Vaccinated&Unvaccinated\\
\hline
Uninfected&\begin{tabular}{c}$14$\\People\end{tabular}&\begin{tabular}{c}$121$\\People\end{tabular}\\
\hline
Asymptomatic&\begin{tabular}{c}$933$\\People\end{tabular}&\begin{tabular}{c}$65$\\People\end{tabular}\\
\hline
Severe Condition&\begin{tabular}{c}$6$\\People\end{tabular}&\begin{tabular}{c}$61$\\People\end{tabular}\\
\hline
Overall&\begin{tabular}{c}$953$\\People\end{tabular}&\begin{tabular}{c}$247$\\People\end{tabular}\\
\hline
\end{tabular}
\caption{Observational data of the clinical study. Here, $1200$ people were free to the vaccine. $953$ people chose to take the vaccine and $247$ people chose to take no vaccine.}
\label{tb2}
\end{table}

Based on the clinical study, the researcher of the vaccine claimed that the vaccine is effective in controlling the severe condition, the number of severe condition patients dropped from $213$ to only $36$.

Now consider the expected difference between benefited and harmed individuals. Recall the benefited individuals include the individual who would be infected in a severe condition if unvaccinated and would be asymptomatic infected if vaccinated, the individual who would be infected in a severe condition if unvaccinated and would be uninfected if vaccinated, and the individual who would be asymptomatic infected if unvaccinated and would be uninfected if vaccinated. The harmed individuals include the individual who would be asymptomatic infected if unvaccinated and would be infected in a severe condition if vaccinated, the individual who would be uninfected if unvaccinated and would be infected in a severe condition if vaccinated, and the individual who would be uninfected if unvaccinated and would be asymptomatic infected if vaccinated. All others are unaffected individuals. In order to maximize the difference between benefited and harmed individuals; therefore, we assign $1$ to benefited individuals, assign $-1$ to harmed individuals, and $0$ to all others in the benefit vector. The objective function (i.e., benefit function) is then\\
\begin{eqnarray*}
f(c)&=&0P({y_1}_{x_1},{y_1}_{x_2}|c)+P({y_1}_{x_1},{y_2}_{x_2}|c)+\\
&&P({y_1}_{x_1},{y_3}_{x_2}|c)-P({y_2}_{x_1},{y_1}_{x_2}|c)+\\
&&0P({y_2}_{x_1},{y_2}_{x_2}|c)+P({y_2}_{x_1},{y_3}_{x_2}|c)-\\
&&-P({y_3}_{x_1},{y_1}_{x_2}|c)-P({y_3}_{x_1},{y_2}_{x_2}|c)+\\
&&0P({y_3}_{x_1},{y_3}_{x_2}|c).
\label{}
\end{eqnarray*}

The experimental data in Table \ref{tb1} provide the following estimates:\\
\begin{eqnarray*}
P({y_1}_{x_1}|c) = 52 / 600 = 0.087\\
P({y_2}_{x_1}|c) = 512 / 600 = 0.853\\
P({y_3}_{x_1}|c) = 36 / 600 = 0.060\\
P({y_1}_{x_2}|c) = 329 / 600 = 0.548\\
P({y_2}_{x_2}|c) = 58 / 600 = 0.097\\
P({y_3}_{x_2}|c) = 213 / 600 = 0.355
\label{}
\end{eqnarray*}

The observational data Table \ref{tb2} provide the following estimates:\\
\begin{eqnarray*}
P(x_1,y_1|c) = 14 / 1200 = 0.012\\
P(x_1,y_2|c) = 933 / 1200 = 0.778\\
P(x_1,y_3|c) = 6 / 1200 = 0.005\\
P(x_2,y_1|c) = 121 / 1200 = 0.101\\
P(x_2,y_2|c) = 65 / 1200 = 0.054\\
P(x_2,y_3|c) = 61 / 1200 = 0.051
\label{}
\end{eqnarray*}

We plug the estimates and the benefit function into Theorem \ref{thm1}, the Algorithm \ref{alg1} returns false (i.e., not identifiable by experimental data). We then plug the estimates and the benefit function into Theorem \ref{thm2} to obtain the bounds
\begin{eqnarray*}
-0.228 \le f(c) \le -0.107
\label{}
\end{eqnarray*}

Thus, the expected difference between benefited and harmed individuals is at most $-0.107$ per individual. We can conclude that the vaccine is ineffective for the virus.

\subsection{Task 2}
The researcher of the clinic study claimed that the individual who would be infected in a severe condition if unvaccinated and would be uninfected if vaccinated and the individual who would be uninfected if unvaccinated and would be infected in a severe condition if vaccinated should be twice important than other individuals. Based on the clinical study, the number of severe condition patients dropped from $213$ to only $36$; therefore, the vaccine should be effective for the virus.

Now consider the expected difference between benefited and harmed individuals. The benefit vector should be the same except assigning $2$ to the individual who would be infected in a severe condition if unvaccinated and would be uninfected if vaccinated and assigning $-2$ to the individual who would be uninfected if unvaccinated and would be infected in a severe condition if vaccinated.

The objective function (i.e., benefit function) is then\\
\begin{eqnarray*}
f(c)&=&0P({y_1}_{x_1},{y_1}_{x_2}|c)+P({y_1}_{x_1},{y_2}_{x_2}|c)+\\
&&2P({y_1}_{x_1},{y_3}_{x_2}|c)-P({y_2}_{x_1},{y_1}_{x_2}|c)+\\
&&0P({y_2}_{x_1},{y_2}_{x_2}|c)+P({y_2}_{x_1},{y_3}_{x_2}|c)-\\
&&-2P({y_3}_{x_1},{y_1}_{x_2}|c)-P({y_3}_{x_1},{y_2}_{x_2}|c)+\\
&&0P({y_3}_{x_1},{y_3}_{x_2}|c).
\label{}
\end{eqnarray*}

We plug the estimates and the benefit function into Theorem \ref{thm1}, the Algorithm \ref{alg1} returns true (i.e., identifiable by experimental data) with value $-0.167$. The benefit function can be simplified as follow:
\begin{eqnarray*}
f(c) &=& 2P({y_1}_{x_1}|c) + P({y_2}_{x_1}|c) -\\
&& - 2P({y_1}_{x_2}|c) - P({y_2}_{x_2}|c)\\
&=&-0.167.
\label{}
\end{eqnarray*}

Thus, the expected difference between benefited and harmed individuals is exactly $-0.167$ per individual. We can conclude that the vaccine is still ineffective for the virus.

\section{Simulated Results}
\label{simres}
In this section, we show the quality of the bounds of the benefit function obtained by Theorem \ref{thm2} using four common benefit vectors.

\begin{figure}
\centering
\includegraphics[width=0.499\textwidth]{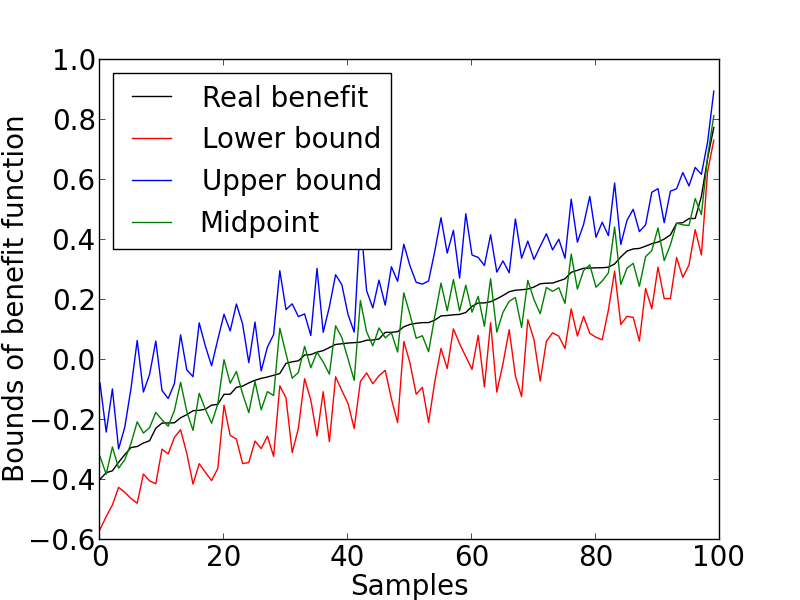}
\caption{Bounds of the benefit function for $100$ sample populations out of $1000$ with the benefit vector $(0,1,1,-1,0,1,-1,-1,0)$.}
\label{res1}
\end{figure}

First, we set $m=2$ (i.e., $X$ has two values) and $n=3$ (i.e., $Y$ has three values). We set the benefit vector to one of the most common ones, $(0,1,1,-1,0,1,-1,-1,0)$, which is to evaluate the expected difference between benefited and harmed individuals. We randomly generated $1000$ populations where each population consists of different fractions of nine response types of individuals. For each population, we then generated sample distributions (observational data and experimental data) compatible with the fractions of response types (see the appendix for the generating algorithm). The advantage of this generating process is that we have the real benefit value (because we know the fractions of the response types) for comparison. Each sample population represents a different instantiate of the population-specific characteristics $C$ in the model. The generating algorithm ensures that the experimental data and observational data satisfy the general relation (i.e., $P(x,y|c)\le P(y_x|c) \le P(x,y|c) + 1 - P(x|c)$). For a sample population $i$, let $[a_i,b_i]$ be the bounds of the benefit function from the proposed theorem. We summarized the following criteria for each population as illustrated in Figure \ref{res1}:
\begin{itemize}
    \item lower bound : $a_i$;
    \item upper bound : $b_i$;
    \item midpoint : $(a_i+b_i)/2$;
    \item real benefit : dot product of the benefit vector and the fractions of response types;
\end{itemize}

From Figure \ref{res1}, it is clear that the proposed bounds obtained from Theorem \ref{thm2} are a good estimation of the real benefit. The lower and upper bounds are closely around the real benefit and the midpoints are almost identified with the real benefit. Besides, the average gap of the bounds, $\frac{\sum(b_i-a_i)}{1000}$, is $0.330$, which is also small compared to the largest possible gap of $6$.

\begin{figure}
\centering
\includegraphics[width=0.499\textwidth]{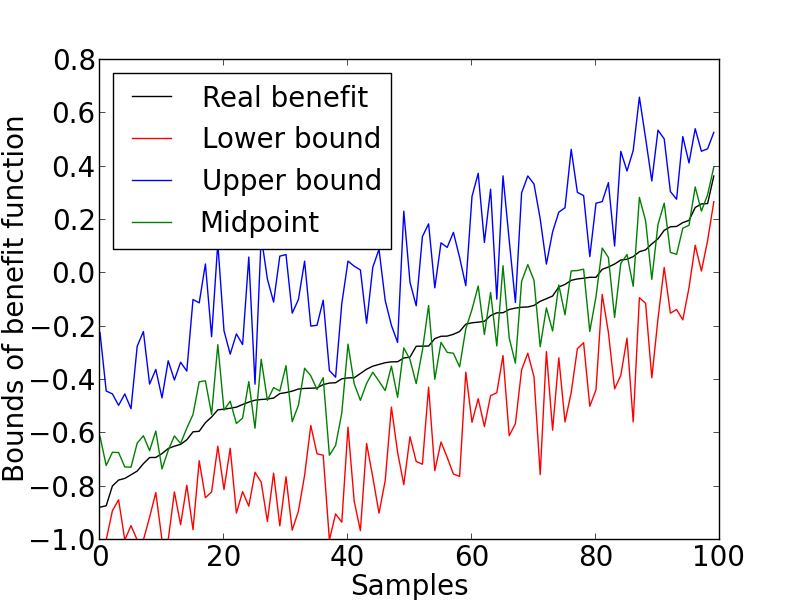}
\caption{Bounds of the benefit function for $100$ sample populations out of $1000$ with the benefit vector $(-1,1,1,-1,-1,1,-1,-1,-1)$.}
\label{res2}
\end{figure}

Second, we set the benefit vector to another common one, $(-1,1,1,-1,-1,1,-1,-1,-1)$, which is to evaluate the expected difference between benefited and unbenefited (i.e., unaffected and harmed) individuals. We again randomly generated $1000$ populations where each population consists of different fractions of nine response types. The data generating process and all other factors remain the same. We summarized the same criteria for each population as illustrated in Figure \ref{res2}.

From Figure \ref{res2}, it is clear that the proposed bounds obtained from Theorem \ref{thm2} are a good estimation of the real benefit. The lower and upper bounds are closely around the real benefit and the midpoints are almost identified with the real benefit. Besides, the average gap of the bounds, $\frac{\sum(b_i-a_i)}{1000}$, is $0.6520$, which is also small compared to the largest possible gap of $9$.

\begin{figure}
\centering
\includegraphics[width=0.499\textwidth]{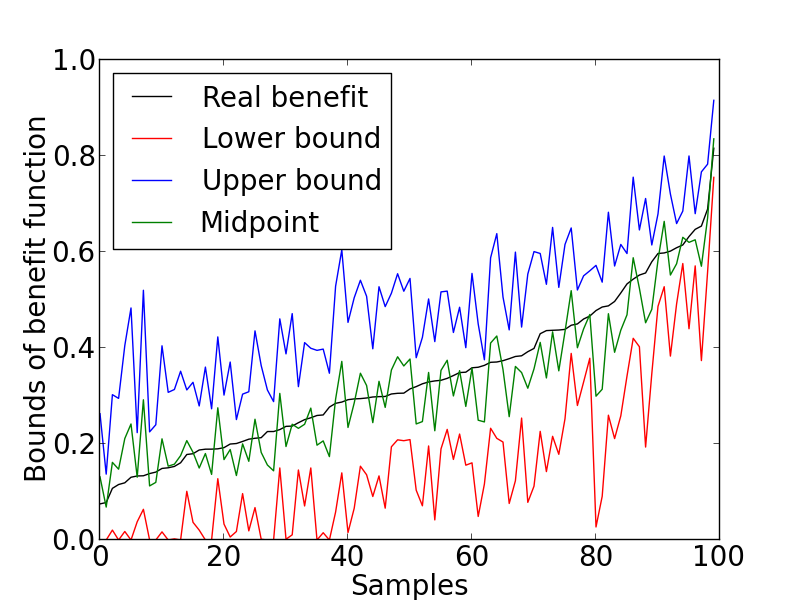}
\caption{Bounds of the benefit function for $100$ sample populations out of $1000$ with the benefit vector $(0,1,1,0,0,1,0,0,0)$.}
\label{res3}
\end{figure}

Third, we set the benefit vector to another common one, $(0,1,1,0,0,1,0,0,0)$, which is to evaluate the expected benefited individuals. We again randomly generated $1000$ populations where each population consists of different fractions of nine response types. The data generating process and all other factors still remain the same. We summarized the same criteria for each population as illustrated in Figure \ref{res3}.

From Figure \ref{res3}, it is clear that the proposed bounds obtained from Theorem \ref{thm2} are a good estimation of the real benefit. The lower and upper bounds are closely around the real benefit and the midpoints are almost identified with the real benefit. Besides, the average gap of the bounds, $\frac{\sum(b_i-a_i)}{1000}$, is $0.3284$, which is also small compared to the largest possible gap of $3$.

\begin{figure}
\centering
\includegraphics[width=0.499\textwidth]{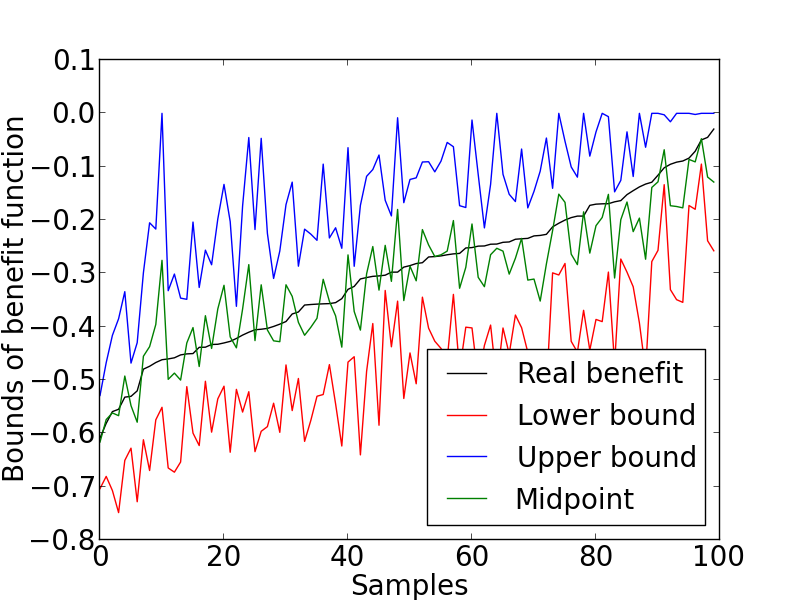}
\caption{Bounds of the benefit function for $100$ sample populations out of $1000$ with the benefit vector $(0,0,0,-1,0,0,-1,-1,0)$.}
\label{res4}
\end{figure}

Lastly, we set the benefit vector to the last common one, $(0,0,0,-1,0,0,-1,-1,0)$, which is to evaluate the expected harmed individuals (we set the benefit vector to $-1$ because we want to minimize the harmed individuals). We again randomly generated $1000$ populations where each population consists of different fractions of nine response types. The data generating process and all other factors still remain the same. We summarized the same criteria for each population as illustrated in Figure \ref{res4}.

From Figure \ref{res4}, it is clear that the proposed bounds obtained from Theorem \ref{thm2} are a good estimation of the real benefit. The lower and upper bounds are closely around the real benefit and the midpoints are almost identified with the real benefit. Besides, the average gap of the bounds, $\frac{\sum(b_i-a_i)}{1000}$, is $0.3266$, which is also small compared to the largest possible gap of $3$.

\section{Discussion}
We have shown that the proposed theorems are a good estimation of the non-binary benefit function using examples and simulated studies. One may concern about the computation complexity of Algorithms \ref{alg1} and \ref{alg2}. They are for sure in exponential time. However, the $m$ and $n$ (i.e., values of $X$ and $Y$) are usually small constant, therefore, we do not need to worry about too much.

\section{Conclusion and Future Work}
We demonstrated the formalization of the general benefit function with nonbinary treatment and effect. We provided the algorithm to compute the bounds of the general benefit function and the algorithm to check whether the benefit function is identifiable with purely experimental data. Examples and simulation results are provided to support the proposed theorems.

Future studies could assess the statistical properties of the proposed bounds. How tight would the bounds be? Does it sufficient to make decisions? Which data, experimental or observational, would affect the bounds more? How would the number of values in treatment and effect affect the quality of the bounds?

Another future direction could be to improve the bounds using covariate information as Li and Pearl \cite{li:pea19-r488} did for the binary benefit function.

\section{Acknowledgements}
This research was supported in parts by grants from the National Science
Foundation [\#IIS-2106908], Office of Naval Research [\#N00014-17-S-12091
and \#N00014-21-1-2351], and Toyota Research Institute of North America
[\#PO-000897].

\bibliography{aaai23.bib}


\clearpage
\newpage
\appendix
\section{Appendix}
\subsection{Proof of Theorems}
\begin{reptheorem}{thm1}
Suppose variables $X$ has $m$ values $x_1,...,x_m$ and $Y$ has $n$ values $y_1,...,y_n$. Then the benefit function $f(c)$ is identifiable if Algorithm \ref{alg1} returns (True, res), and res is the value of the benefit function.\\
\begin{eqnarray*}
f(c) =&& \alpha_1 P({y_1}_{x_1},{y_1}_{x_2},...,{y_1}_{x_m}|c)+\nonumber\\
&&\alpha_2 P({y_1}_{x_1},{y_1}_{x_2},...,{y_2}_{x_m}|c)+... \nonumber \\
&&\alpha_n P({y_1}_{x_1},{y_1}_{x_2},...,{y_n}_{x_m}|c)+... \nonumber \\
&&\alpha_{n^{m-1}+1} P({y_2}_{x_1},{y_1}_{x_2},...,{y_1}_{x_m}|c)+...\nonumber\\
&&\alpha_{n^m} P({y_n}_{x_1},{y_n}_{x_2},...,{y_n}_{x_m}|c).
\end{eqnarray*}
\begin{proof}
The proof is simple.\\
Lines $1$ to $12$ in Algorithm \ref{alg1} simply check whether the given benefit function $f$ (encoded as $a$ in the algorithm) is identifiable.\\
On line $24$ of the algorithm, we recursively call on another benefit function $f'$ (encoded as $na$ in the algorithm).\\
Now lets consider how we obtain $f'$.\\
if there exist $n^{m-1}$ terms in $a$ and $r$, s.t., $a[j_1][r]=...=a[j_{n^{m-1}}][r]$, these $n^{m-1}$ terms are $P(...,{y_{a[j_1][r]}}_{x_{r-1}},...|c)$, and the sum of these $n^{m-1}$ terms is equal to $P({y_{a[j_1][r]}}_{x_{r-1}}|c)$.\\
We obtain $f'$ by eliminating kth of the $n^{m-1}$ terms in $f$, replacing kth term by other $n^{m-1}-1$ terms and their sum $P({y_{a[j_1][r]}}_{x_{r-1}}|c)$.\\
Therefore, $f=f'+nc$ where $nc=a[j_k][1]*P({y_{a[j_1][r]}}_{x_{r-1}}|c)$.
Thus, $f$ is identifiable if and only if $f'$ is identifiable and $f = f' + nc$.
\end{proof}
\end{reptheorem}

\begin{reptheorem}{thm2}
Suppose variables $X$ has $m$ values $x_1,...,x_m$ and $Y$ has $n$ values $y_1,...,y_n$. Then the bounds of the benefit function $f(c)$ is obtained by Algorithm \ref{alg2}.\\
\begin{eqnarray*}
f(c) =&& \alpha_1 P({y_1}_{x_1},{y_1}_{x_2},...,{y_1}_{x_m}|c)+\nonumber\\
&&\alpha_2 P({y_1}_{x_1},{y_1}_{x_2},...,{y_2}_{x_m}|c)+... \nonumber \\
&&\alpha_n P({y_1}_{x_1},{y_1}_{x_2},...,{y_n}_{x_m}|c)+... \nonumber \\
&&\alpha_{n^{m-1}+1} P({y_2}_{x_1},{y_1}_{x_2},...,{y_1}_{x_m}|c)+...\nonumber\\
&&\alpha_{n^m} P({y_n}_{x_1},{y_n}_{x_2},...,{y_n}_{x_m}|c).
\end{eqnarray*}
\begin{proof}
Similarly to Theorem \ref{thm1},\\
lines $1$ to $12$ in Algorithm \ref{alg2} simply compute the bounds of the given benefit function $f$ (encoded as $a$ in the algorithm).\\
On line $24$ of the algorithm, we recursively call on another benefit function $f'$ (encoded as $na$ in the algorithm).\\
Now lets consider how we obtain $f'$.\\
if there exist $n^{m-1}$ terms in $a$ and $r$, s.t., $a[j_1][r]=...=a[j_{n^{m-1}}][r]$, these $n^{m-1}$ terms are $P(...,{y_{a[j_1][r]}}_{x_{r-1}},...|c)$, and the sum of these $n^{m-1}$ terms is equal to $P({y_{a[j_1][r]}}_{x_{r-1}}|c)$.\\
We obtain $f'$ by eliminating kth of the $n^{m-1}$ terms in $f$, replacing kth term by other $n^{m-1}-1$ terms and their sum $P({y_{a[j_1][r]}}_{x_{r-1}}|c)$.\\
Therefore, $f=f'+nc$ where $nc=a[j_k][1]*P({y_{a[j_1][r]}}_{x_{r-1}}|c)$.
Thus, the bounds of $f'+nc$ is the bounds of $f$.
\end{proof}
\end{reptheorem}

\subsection{Li-Pearl's Bounds of Probabilities of Causation}
The input, $lb,ub$, in Algorithm \ref{alg2} depends on the bounds of probabilities of causation. The bounds of probabilities of causation recently proposed by Li and Pearl \cite{li:pea-r516} is not conditional $C$. However, nothing is changed if conditioning on a variable $C$ that is not affected by $X$. We listed the conditional version of the eight theorems proposed by Li and Pearl. The proof of eight theorems is exactly the same, except every probability should be conditioned on $C$. 
\begin{theorem}
\label{thm4}
Suppose variable $X$ has $m$ values $x_1,...,x_m$ and $Y$ has $n$ values $y_1,...,y_n$, and variable $C$ is not affected by $X$, then the probability of causation $P({y_i}_{x_j}, y_i|c)$, where $1 \le i \le n, 1 \le j \le m$, is bounded as following:
\begin{eqnarray*}
\max \left \{
\begin{array}{cc}
P(x_j, y_i|c), \\
P({y_i}_{x_j}|c) + P(y_i|c) - 1 \\
\end{array}
\right \}
\le P({y_i}_{x_j}, y_i|c)
\label{t4e1}
\end{eqnarray*}
\begin{eqnarray*}
P({y_i}_{x_j}, y_i|c) \le \min \left \{
\begin{array}{cc}
 P({y_i}_{x_j}|c), \\
 P(y_i|c) \\
\end{array} 
\right \}
\label{t4e2}
\end{eqnarray*}
\end{theorem}

\begin{theorem}
\label{thm5}
Suppose variable $X$ has $m$ values $x_1,...,x_m$ and $Y$ has $n$ values $y_1,...,y_n$, and variable $C$ is not affected by $X$, then the probability of causation $P({y_i}_{x_j}, y_k|c)$, where $1 \le i,k \le n, 1 \le j \le m, i\ne k$, is bounded as following:
\begin{eqnarray*}
\max \left \{
\begin{array}{cc}
0, \\
P({y_i}_{x_j}|c) + P(y_k|c) - 1, \\
\sum_{1\le p\le m,p\ne j}\max \left \{
\begin{array}{cc}
0, \\
P({y_i}_{x_j}|c)\\
+ P(x_p,y_k|c) \\
- 1 + P(x_j|c)\\
- P(x_j,y_i|c) \\
\end{array}
\right \}
\end{array}
\right \}\nonumber\\
\le P({y_i}_{x_j}, y_k|c)
\label{t5e1}
\end{eqnarray*}
\begin{eqnarray*}
P({y_i}_{x_j}, y_k|c) \le \min \left \{
\begin{array}{cc}
 P({y_i}_{x_j}|c) - P(x_j, y_i|c), \\
 P(y_k|c) - P(y_k, x_j|c) \\
\end{array} 
\right \}
\label{t5e2}
\end{eqnarray*}
\end{theorem}

\begin{theorem}
\label{thm6}
Suppose variable $X$ has $m$ values $x_1,...,x_m$ and $Y$ has $n$ values $y_1,...,y_n$, and variable $C$ is not affected by $X$, then the probability of causation $P({y_i}_{x_j}, x_k|c)$, where $1 \le i \le n, 1 \le j,k \le m, j\ne k$, is bounded as following:
\begin{eqnarray*}
\max \left \{
\begin{array}{cc}
0, \\
P({y_i}_{x_j}|c) - P(x_j, y_i|c)\\
- 1 + P(x_j|c) + P(x_k|c) \\
\end{array}
\right \}
\le P({y_i}_{x_j}, x_k|c)
\label{t6e1}
\end{eqnarray*}
\begin{eqnarray*}
P({y_i}_{x_j}, x_k|c) \le \min \left \{
\begin{array}{cc}
 P({y_i}_{x_j}|c) - P(x_j, y_i|c), \\
 P(x_k|c) \\
\end{array} 
\right \}
\label{t6e2}
\end{eqnarray*}
\end{theorem}

\begin{theorem}
\label{thm7}
Suppose variable $X$ has $m$ values $x_1,...,x_m$ and $Y$ has $n$ values $y_1,...,y_n$, and variable $C$ is not affected by $X$, then the probability of causation $P({y_i}_{x_j}, y_k, x_p|c)$, where $1 \le i,k \le n, 1 \le j,p \le m, j\ne p$, is bounded as following:
\begin{eqnarray*}
\max \left \{
\begin{array}{cc}
0, \\
 P({y_i}_{x_j}|c) + P(x_p, y_k|c)\\
- 1 + P(x_j|c) - P(x_j, y_i|c) \\
\end{array}
\right \}
\le P({y_i}_{x_j}, y_k, x_p|c)
\label{t7e1}
\end{eqnarray*}
\begin{eqnarray*}
P({y_i}_{x_j}, y_k, x_p|c) \le \min \left \{
\begin{array}{cc}
 P({y_i}_{x_j}|c) - P(x_j, y_i|c), \\
 P(x_p, y_k|c) \\
\end{array} 
\right \}
\label{t7e2}
\end{eqnarray*}
\end{theorem}


\begin{theorem}
\label{thm8}
Suppose variable $X$ has $m$ values $x_1,...,x_m$ and $Y$ has $n$ values $y_1,...,y_n$, and variable $C$ is not affected by $X$, then the probability of causation $P({y_{i_1}}_{x_{j_1}},...,{y_{i_k}}_{x_{j_k}}|c)$, where $1 \le i_1,...,i_k \le n, 1 \le j_1,...,j_k \le m, j_1\ne ... \ne j_k$, is bounded as following:
\begin{eqnarray*}
\max \left \{
\begin{array}{cc}
0, \\
\\
\sum_{1\le t\le k}P({y_{i_t}}_{x_{j_t}}|c) - k + 1, \\
\\
\max_{1\le t \le k} (LB(P({y_{i_1}}_{x_{j_1}},...,{y_{i_{t-1}}}_{x_{j_{t-1}}},\\
{y_{i_{t+1}}}_{x_{j_{t+1}}},...,{y_{i_k}}_{x_{j_k}}|c))\\
+ P({y_{i_t}}_{x_{i_t}}|c) - 1),\\
\\
\sum_{1\le p\le m, s.t., \exists r, 1\le r\le k, p=j_r }\\ LB(P({y_{i_1}}_{x_{j_1}},...,{y_{i_{r-1}}}_{x_{j_{r-1}}},\\
{y_{i_{r+1}}}_{x_{j_{r+1}}},...,{y_{i_k}}_{x_{j_k}}, x_{j_r}, y_{i_r}|c)) + \\
\sum_{1\le p\le m, s.t., p \ne j_1 \ne ...\ne j_k}\\ LB(P({y_{i_1}}_{x_{j_1}},...,{y_{i_k}}_{x_{j_k}}, x_p|c))\\
\end{array}
\right \}\nonumber\\
\le P({y_{i_1}}_{x_{j_1}},...,{y_{i_k}}_{x_{j_k}}|c)
\label{t8e1}
\end{eqnarray*}
\begin{eqnarray*}
P({y_{i_1}}_{x_{j_1}},...,{y_{i_k}}_{x_{j_k}}|c) \le \nonumber\\
\min \left \{
\begin{array}{cc}
 \min_{1\le t\le k} P({y_{i_t}}_{x_{j_t}}|c), \\
 \\
 \min_{1\le t \le k} UB(P({y_{i_1}}_{x_{j_1}},...,{y_{i_{t-1}}}_{x_{j_{t-1}}},\\
{y_{i_{t+1}}}_{x_{j_{t+1}}},...,{y_{i_k}}_{x_{j_k}}|c)),\\
 \\
\sum_{1\le p\le m, s.t., \exists r, 1\le r\le k, p=j_r}\\ UB(P({y_{i_1}}_{x_{j_1}},...,{y_{i_{r-1}}}_{x_{j_{r-1}}},\\
{y_{i_{r+1}}}_{x_{j_{r+1}}},...,{y_{i_k}}_{x_{j_k}}, x_{j_r}, y_{i_r}|c)) + \\
\sum_{1\le p\le m, s.t., p \ne j_1 \ne ...\ne j_k}\\ UB(P({y_{i_1}}_{x_{j_1}},...,{y_{i_k}}_{x_{j_k}}, x_p|c))\\
\end{array} 
\right \}
\label{t8e2}
\end{eqnarray*}
where,\\
LB$(f)$ denotes the lower bound of a function $f$ and UB$(f)$ denotes the upper bound of a function $f$. The bounds of $P({y_{i_1}}_{x_{j_1}},...,{y_{i_{r-1}}}_{x_{j_{r-1}}},{y_{i_{r+1}}}_{x_{j_{r+1}}},...,{y_{i_k}}_{x_{j_k}}, x_{j_r}, y_{j_r}|c)$ are given by Theorem \ref{thm7} or \ref{thm11}, the bounds of $P({y_{i_1}}_{x_{j_1}},...,{y_{i_k}}_{x_{j_k}}, x_p|c)$ are given by Theorem \ref{thm6} or \ref{thm9}, and the bounds of $P({y_{i_1}}_{x_{j_1}},...,{y_{i_{t-1}}}_{x_{j_{t-1}}},{y_{i_{t+1}}}_{x_{j_{t+1}}},...,{y_{i_k}}_{x_{j_k}}|c)$ are given by Theorem \ref{thm8} or experimental data if $k=2$.
\end{theorem}

\begin{theorem}
\label{thm9}
Suppose variable $X$ has $m$ values $x_1,...,x_m$ and $Y$ has $n$ values $y_1,...,y_n$, and variable $C$ is not affected by $X$, then the probability of causation $P({y_{i_1}}_{x_{j_1}},...,{y_{i_k}}_{x_{j_k}},x_p|c)$, where $1 \le i_1,...,i_k \le n, 1 \le j_1,...,j_k,p \le m, j_1\ne ... \ne j_k \ne p$, is bounded as following:
\begin{eqnarray*}
\max \left \{
\begin{array}{cc}
0, \\
\\
\sum_{1\le t\le k}P({y_{i_t}}_{x_{j_t}}|c) +P(x_p|c) - k, \\
\\
\max_{1\le t \le k} (LB(P({y_{i_1}}_{x_{j_1}},...,{y_{i_{t-1}}}_{x_{j_{t-1}}},\\
{y_{i_{t+1}}}_{x_{j_{t+1}}},...,{y_{i_k}}_{x_{j_k}}|c))\\
+ LB(P({y_{i_t}}_{x_{i_t}},x_p|c)) - 1)\\
\end{array}
\right \}\nonumber\\
\le P({y_{i_1}}_{x_{j_1}},...,{y_{i_k}}_{x_{j_k}},x_p|c)
\label{t9e1}
\end{eqnarray*}
\begin{eqnarray*}
P({y_{i_1}}_{x_{j_1}},...,{y_{i_k}}_{x_{j_k}},x_p|c) \le \nonumber\\
\min \left \{
\begin{array}{cc}
 \min_{1\le t\le k} P({y_{i_t}}_{x_{j_t}}|c), \\
 \\
 P(x_p|c),\\
 \\
\min_{1\le t \le k} UB(P({y_{i_1}}_{x_{j_1}},...,{y_{i_{t-1}}}_{x_{j_{t-1}}},\\
{y_{i_{t+1}}}_{x_{j_{t+1}}},...,{y_{i_k}}_{x_{j_k}}|c)),\\
\\
\min_{1\le t \le k} UB(P({y_{i_t}}_{x_{i_t}},x_p|c))\\
\end{array} 
\right \}
\label{t9e2}
\end{eqnarray*}
where,\\
LB$(f)$ denotes the lower bound of a function $f$ and UB$(f)$ denotes the upper bound of a function $f$. The bounds of $P({y_{i_1}}_{x_{j_1}},...,{y_{i_{t-1}}}_{x_{j_{t-1}}},{y_{i_{t+1}}}_{x_{j_{t+1}}},...,{y_{i_k}}_{x_{j_k}}|c)$ are given by Theorem \ref{thm8} or experimental data if $k=2$ and the bounds of $P({y_{i_t}}_{x_{i_t}},x_p|c)$ are given by Theorem \ref{thm6}.
\end{theorem}

\begin{theorem}
\label{thm10}
Suppose variable $X$ has $m$ values $x_1,...,x_m$ and $Y$ has $n$ values $y_1,...,y_n$, and variable $C$ is not affected by $X$, then the probability of causation $P({y_{i_1}}_{x_{j_1}},...,{y_{i_k}}_{x_{j_k}},y_q|c)$, where $1 \le i_1,...,i_k,q \le n, 1 \le j_1,...,j_k \le m, j_1\ne ... \ne j_k$, is bounded as following:
\begin{eqnarray*}
\max \left \{
\begin{array}{cc}
0,\\
\\
\sum_{1\le t\le k}P({y_{i_t}}_{x_{j_t}}|c) +P(y_q|c) - k, \\
\\
\max_{1\le t \le k} (LB(P({y_{i_1}}_{x_{j_1}},...,{y_{i_{t-1}}}_{x_{j_{t-1}}},\\
{y_{i_{t+1}}}_{x_{j_{t+1}}},...,{y_{i_k}}_{x_{j_k}}|c))\\
+ LB(P({y_{i_t}}_{x_{i_t}},y_q|c)) - 1),\\
\\
\sum_{1\le p\le m, \exists r, 1\le r\le k, p=j_r, q=i_r}\\ LB(P({y_{i_1}}_{x_{j_1}},...,{y_{i_{r-1}}}_{x_{j_{r-1}}},\\
{y_{i_{r+1}}}_{x_{j_{r+1}}},...,{y_{i_k}}_{x_{j_k}}, x_{j_r}, y_{i_r}|c)) + \\
\sum_{1\le p\le m, s.t., p \ne j_1 \ne ...\ne j_k}\\ LB(P({y_{i_1}}_{x_{j_1}},...,{y_{i_k}}_{x_{j_k}}, x_p, y_q|c))\\
\end{array}
\right \}\nonumber\\
\le P({y_{i_1}}_{x_{j_1}},...,{y_{i_k}}_{x_{j_k}}, y_q|c)
\label{t10e1}
\end{eqnarray*}
\begin{eqnarray*}
P({y_{i_1}}_{x_{j_1}},...,{y_{i_k}}_{x_{j_k}},y_q|c) \le \nonumber\\
\min \left \{
\begin{array}{cc}
 \min_{1\le t\le k} P({y_{i_t}}_{x_{j_t}}|c), \\
 \\
 P(y_q|c),\\
 \\
\min_{1\le t \le k} UB(P({y_{i_1}}_{x_{j_1}},...,{y_{i_{t-1}}}_{x_{j_{t-1}}},\\
{y_{i_{t+1}}}_{x_{j_{t+1}}},...,{y_{i_k}}_{x_{j_k}}|c)),\\
\\
\min_{1\le t \le k} UB(P({y_{i_t}}_{x_{i_t}},y_q|c)),\\
\\
\sum_{1\le p\le m, s.t., \exists r, 1\le r\le k, p=j_r, q=i_r}\\ UB(P({y_{i_1}}_{x_{j_1}},...,{y_{i_{r-1}}}_{x_{j_{r-1}}},\\
{y_{i_{r+1}}}_{x_{j_{r+1}}},...,{y_{i_k}}_{x_{j_k}}, x_{j_r}, y_{i_r}|c)) + \\
\sum_{1\le p\le m, s.t., p \ne j_1 \ne ...\ne j_k}\\ UB(P({y_{i_1}}_{x_{j_1}},...,{y_{i_k}}_{x_{j_k}}, x_p, y_q|c))\\
\end{array} 
\right \}
\label{t10e2}
\end{eqnarray*}
where,\\
LB$(f)$ denotes the lower bound of a function $f$ and UB$(f)$ denotes the upper bound of a function $f$. The bounds of $P({y_{i_1}}_{x_{j_1}},...,{y_{i_{r-1}}}_{x_{j_{r-1}}},{y_{i_{r+1}}}_{x_{j_{r+1}}},...,{y_{i_k}}_{x_{j_k}}, x_{j_r}, y_{j_r}|c)$, $P({y_{i_1}}_{x_{j_1}},...,{y_{i_k}}_{x_{j_k}}, x_p, y_q|c)$ are given by Theorem \ref{thm7} or \ref{thm11}, the bounds of $P({y_{i_1}}_{x_{j_1}},...,{y_{i_{t-1}}}_{x_{j_{t-1}}}, {y_{i_{t+1}}}_{x_{j_{t+1}}},...,{y_{i_k}}_{x_{j_k}}|c)$ are given by Theorem \ref{thm8} or experimental data if $k=2$, and the bounds of $P({y_{i_t}}_{x_{i_t}},y_q|c)$ are given by Theorem \ref{thm4} or \ref{thm5}.
\end{theorem}

\begin{theorem}
\label{thm11}
Suppose variable $X$ has $m$ values $x_1,...,x_m$ and $Y$ has $n$ values $y_1,...,y_n$, and variable $C$ is not affected by $X$, then the probability of causation $P({y_{i_1}}_{x_{j_1}},...,{y_{i_k}}_{x_{j_k}},x_p,y_q|c)$, where $1 \le i_1,...,i_k,q \le n, 1 \le j_1,...,j_k,p \le m, j_1\ne ... \ne j_k \ne p$, is bounded as following:
\begin{eqnarray*}
\max \left \{
\begin{array}{cc}
0, \\
\\
\sum_{1\le t\le k}P({y_{i_t}}_{x_{j_t}}|c) +P(x_p,y_q|c) - k, \\
\\
\max_{1\le t \le k} (LB(P({y_{i_1}}_{x_{j_1}},...,{y_{i_{t-1}}}_{x_{j_{t-1}}},\\
{y_{i_{t+1}}}_{x_{j_{t+1}}},...,{y_{i_k}}_{x_{j_k}}|c))\\
+ LB(P({y_{i_t}}_{x_{i_t}},x_p,y_q|c)) - 1)\\
\end{array}
\right \}\nonumber\\
\le P({y_{i_1}}_{x_{j_1}},...,{y_{i_k}}_{x_{j_k}},x_p,y_q|c)
\label{t11e1}
\end{eqnarray*}
\begin{eqnarray*}
P({y_{i_1}}_{x_{j_1}},...,{y_{i_k}}_{x_{j_k}},x_p,y_q|c) \le \nonumber\\
\min \left \{
\begin{array}{cc}
 \min_{1\le t\le k} P({y_{i_t}}_{x_{j_t}}|c), \\
 \\
 P(x_p,y_q|c),\\
 \\
\min_{1\le t \le k} UB(P({y_{i_1}}_{x_{j_1}},...,{y_{i_{t-1}}}_{x_{j_{t-1}}},\\
{y_{i_{t+1}}}_{x_{j_{t+1}}},...,{y_{i_k}}_{x_{j_k}}|c)),\\
\\
\min_{1\le t \le k} UB(P({y_{i_t}}_{x_{i_t}},x_p,y_q|c))\\
\end{array} 
\right \}
\label{t11e2}
\end{eqnarray*}
where,\\
LB$(f)$ denotes the lower bound of a function $f$ and UB$(f)$ denotes the upper bound of a function $f$. The bounds of $P({y_{i_1}}_{x_{j_1}},...,{y_{i_{t-1}}}_{x_{j_{t-1}}}, {y_{i_{t+1}}}_{x_{j_{t+1}}},...,{y_{i_k}}_{x_{j_k}}|c)$ are given by Theorem \ref{thm8} or experimental data if $k=2$ and the bounds of $P({y_{i_t}}_{x_{i_t}},x_p,y_q|c)$ are given by Theorem \ref{thm7}.
\end{theorem}

\subsection{Calculation in the Example}
\subsubsection{Task 1}
First by Theorem \ref{thm8}, we have,
\begin{eqnarray*}
0 \le P({y_1}_{x_1},{y_1}_{x_2}|c) \le 0.087,\\
0 \le P({y_1}_{x_1},{y_2}_{x_2}|c) \le 0.066,\\
0 \le P({y_1}_{x_1},{y_3}_{x_2}|c) \le 0.063,\\
0.431 \le P({y_2}_{x_1},{y_1}_{x_2}|c) \le 0.523,\\
0.026 \le P({y_2}_{x_1},{y_2}_{x_2}|c) \le 0.097,\\
0.287 \le P({y_2}_{x_1},{y_3}_{x_2}|c) \le 0.355,\\
0 \le P({y_3}_{x_1},{y_1}_{x_2}|c) \le 0.060,\\
0 \le P({y_3}_{x_1},{y_2}_{x_2}|c) \le 0.059,\\
0 \le P({y_3}_{x_1},{y_3}_{x_2}|c) \le 0.056.
\end{eqnarray*}
By Algorithm \ref{alg2}, the lower bound came from the following steps,
\begin{eqnarray*}
f(c)&=&0P({y_1}_{x_1},{y_1}_{x_2}|c)+P({y_1}_{x_1},{y_2}_{x_2}|c)+\\
&&P({y_1}_{x_1},{y_3}_{x_2}|c)-P({y_2}_{x_1},{y_1}_{x_2}|c)+\\
&&0P({y_2}_{x_1},{y_2}_{x_2}|c)+P({y_2}_{x_1},{y_3}_{x_2}|c)-\\
&&-P({y_3}_{x_1},{y_1}_{x_2}|c)-P({y_3}_{x_1},{y_2}_{x_2}|c)+\\
&&0P({y_3}_{x_1},{y_3}_{x_2}|c)\\
&=&P({y_1}_{x_1},{y_2}_{x_2}|c)+\\
&&P({y_1}_{x_1},{y_3}_{x_2}|c)-P({y_2}_{x_1},{y_1}_{x_2}|c)+\\
&&0P({y_2}_{x_1},{y_2}_{x_2}|c)+P({y_2}_{x_1},{y_3}_{x_2}|c)-\\
&&-P({y_3}_{x_1},{y_1}_{x_2}|c)-P({y_3}_{x_1},{y_2}_{x_2}|c)+\\
&&0P({y_3}_{x_1},{y_3}_{x_2}|c)\\
&=&P({y_1}_{x_1},{y_2}_{x_2}|c)+\\
&&P({y_1}_{x_1},{y_3}_{x_2}|c)-P({y_2}_{x_1},{y_1}_{x_2}|c)+\\
&&P({y_2}_{x_1},{y_3}_{x_2}|c)-\\
&&-P({y_3}_{x_1},{y_1}_{x_2}|c)-P({y_3}_{x_1},{y_2}_{x_2}|c)+\\
&&0P({y_3}_{x_1},{y_3}_{x_2}|c)\\
&=&P({y_1}_{x_1},{y_2}_{x_2}|c)-P({y_2}_{x_1},{y_1}_{x_2}|c)+\\
&&0P({y_2}_{x_1},{y_3}_{x_2}|c)-\\
&&-P({y_3}_{x_1},{y_1}_{x_2}|c)-P({y_3}_{x_1},{y_2}_{x_2}|c)-\\
&&-P({y_3}_{x_1},{y_3}_{x_2}|c)+P({y_3}_{x_2}|c)\\
&=&P({y_1}_{x_1},{y_2}_{x_2}|c)-P({y_2}_{x_1},{y_1}_{x_2}|c)+\\
&&0P({y_2}_{x_1},{y_3}_{x_2}|c)+0P({y_3}_{x_1},{y_2}_{x_2}|c)+\\
&&0P({y_3}_{x_1},{y_3}_{x_2}|c)+P({y_3}_{x_2}|c)-P({y_3}_{x_1}|c)\\
&\ge&LB(P({y_1}_{x_1},{y_2}_{x_2}|c))-UB(P({y_2}_{x_1},{y_1}_{x_2}|c))+\\
&&+0+0+0+213/600-36/600\\
&=&0-0.523+0.295\\
&=&-0.228.
\label{}
\end{eqnarray*}
and the upper bounds came from the following steps,
\begin{eqnarray*}
f(c)&=&0P({y_1}_{x_1},{y_1}_{x_2}|c)+P({y_1}_{x_1},{y_2}_{x_2}|c)+\\
&&P({y_1}_{x_1},{y_3}_{x_2}|c)-P({y_2}_{x_1},{y_1}_{x_2}|c)+\\
&&0P({y_2}_{x_1},{y_2}_{x_2}|c)+P({y_2}_{x_1},{y_3}_{x_2}|c)-\\
&&-P({y_3}_{x_1},{y_1}_{x_2}|c)-P({y_3}_{x_1},{y_2}_{x_2}|c)+\\
&&0P({y_3}_{x_1},{y_3}_{x_2}|c)\\
&=&P({y_1}_{x_1},{y_2}_{x_2}|c)+\\
&&P({y_1}_{x_1},{y_3}_{x_2}|c)-P({y_2}_{x_1},{y_1}_{x_2}|c)+\\
&&0P({y_2}_{x_1},{y_2}_{x_2}|c)+P({y_2}_{x_1},{y_3}_{x_2}|c)-\\
&&-P({y_3}_{x_1},{y_1}_{x_2}|c)-P({y_3}_{x_1},{y_2}_{x_2}|c)+\\
&&0P({y_3}_{x_1},{y_3}_{x_2}|c)\\
&=&P({y_1}_{x_1},{y_3}_{x_2}|c)-P({y_2}_{x_1},{y_1}_{x_2}|c)-\\
&&-P({y_2}_{x_1},{y_2}_{x_2}|c)+P({y_2}_{x_1},{y_3}_{x_2}|c)-\\
&&-P({y_3}_{x_1},{y_1}_{x_2}|c)-2P({y_3}_{x_1},{y_2}_{x_2}|c)+\\
&&0P({y_3}_{x_1},{y_3}_{x_2}|c)+P({y_2}_{x_2}|c)\\
&=&-P({y_2}_{x_1},{y_1}_{x_2}|c)-\\
&&-P({y_2}_{x_1},{y_2}_{x_2}|c)+0P({y_2}_{x_1},{y_3}_{x_2}|c)-\\
&&-P({y_3}_{x_1},{y_1}_{x_2}|c)-2P({y_3}_{x_1},{y_2}_{x_2}|c)-\\
&&-P({y_3}_{x_1},{y_3}_{x_2}|c)+P({y_2}_{x_2}|c)+P({y_3}_{x_2}|c)\\
&=&0P({y_2}_{x_1},{y_2}_{x_2}|c)+1P({y_2}_{x_1},{y_3}_{x_2}|c)-\\
&&-P({y_3}_{x_1},{y_2}_{x_2}|c)+0P({y_3}_{x_1},{y_3}_{x_2}|c)+\\
&&+P({y_2}_{x_2}|c)+P({y_3}_{x_2}|c)-P({y_2}_{x_1}|c)-P({y_3}_{x_1}|c)\\
&\le&0+UB(P({y_2}_{x_1},{y_3}_{x_2}|c))-\\
&&-LB(P({y_3}_{x_1},{y_2}_{x_2}|c))+0+\\
&&+58/600+213/600-512/600-36/600\\
&=&0+0.355-0+0-0.462\\
&=&-0.107.
\label{}
\end{eqnarray*}
Thus,
\begin{eqnarray*}
-0.228 \le f(c)\le -0.107.\\
\end{eqnarray*}
\newpage
\subsubsection{Task 2}
By Algorithm \ref{alg1}, the result came from the following steps,
\begin{eqnarray*}
f(c)&=&0P({y_1}_{x_1},{y_1}_{x_2}|c)+P({y_1}_{x_1},{y_2}_{x_2}|c)+\\
&&2P({y_1}_{x_1},{y_3}_{x_2}|c)-P({y_2}_{x_1},{y_1}_{x_2}|c)+\\
&&0P({y_2}_{x_1},{y_2}_{x_2}|c)+P({y_2}_{x_1},{y_3}_{x_2}|c)-\\
&&-2P({y_3}_{x_1},{y_1}_{x_2}|c)-P({y_3}_{x_1},{y_2}_{x_2}|c)+\\
&&0P({y_3}_{x_1},{y_3}_{x_2}|c)\\
&=&P({y_1}_{x_1},{y_2}_{x_2}|c)+\\
&&2P({y_1}_{x_1},{y_3}_{x_2}|c)-P({y_2}_{x_1},{y_1}_{x_2}|c)+\\
&&0P({y_2}_{x_1},{y_2}_{x_2}|c)+P({y_2}_{x_1},{y_3}_{x_2}|c)-\\
&&-2P({y_3}_{x_1},{y_1}_{x_2}|c)-P({y_3}_{x_1},{y_2}_{x_2}|c)+\\
&&0P({y_3}_{x_1},{y_3}_{x_2}|c)\\
&=&2P({y_1}_{x_1},{y_3}_{x_2}|c)-P({y_2}_{x_1},{y_1}_{x_2}|c)-\\
&&-P({y_2}_{x_1},{y_2}_{x_2}|c)+P({y_2}_{x_1},{y_3}_{x_2}|c)-\\
&&-2P({y_3}_{x_1},{y_1}_{x_2}|c)-2P({y_3}_{x_1},{y_2}_{x_2}|c)+\\
&&0P({y_3}_{x_1},{y_3}_{x_2}|c) + P({y_2}_{x_2}|c)\\
&=&-P({y_2}_{x_1},{y_1}_{x_2}|c)-\\
&&-P({y_2}_{x_1},{y_2}_{x_2}|c)-P({y_2}_{x_1},{y_3}_{x_2}|c)-\\
&&-2P({y_3}_{x_1},{y_1}_{x_2}|c)-2P({y_3}_{x_1},{y_2}_{x_2}|c)-\\
&&-2P({y_3}_{x_1},{y_3}_{x_2}|c) + P({y_2}_{x_2}|c) + 2P({y_3}_{x_2}|c)\\
&=&0P({y_2}_{x_1},{y_2}_{x_2}|c)+0P({y_2}_{x_1},{y_3}_{x_2}|c)-\\
&&-2P({y_3}_{x_1},{y_1}_{x_2}|c)-2P({y_3}_{x_1},{y_2}_{x_2}|c)-\\
&&-2P({y_3}_{x_1},{y_3}_{x_2}|c) +
\\&&+ P({y_2}_{x_2}|c) + 2P({y_3}_{x_2}|c)-P({y_2}_{x_1}|c)\\
&=&0P({y_2}_{x_1},{y_2}_{x_2}|c)+0P({y_2}_{x_1},{y_3}_{x_2}|c)+\\
&&+0P({y_3}_{x_1},{y_2}_{x_2}|c)+0P({y_3}_{x_1},{y_3}_{x_2}|c)+\\
&&P({y_2}_{x_2}|c) + 2P({y_3}_{x_2}|c)-\\
&&-P({y_2}_{x_1}|c)-2P({y_3}_{x_1}|c)\\
&=&58/600+426/600-512/600-72/600\\
&=&-0.167.
\label{}
\end{eqnarray*}

\subsection{Distribution Generating Algorithm}
Here, the sample distribution generating algorithm in the simulated studies is presented. It generated both experimental and observational data compatible with the fractions of response types of individuals. The data satisfy the general relation between experimental and observational data. Note that all four simulated studies shared the same distribution generating algorithm but with different benefit vectors.

\begin{algorithm}[tb]
\caption{Generate sample distributions for simulated studies}
\label{alg3}
\textbf{Input}: $num$, number of samples needed.\\
\textbf{Output}: $num$ sample distributions (observational data and experimental data).
\begin{algorithmic}[1] 
\STATE $count = 0$;
\WHILE {$count<num$}
    \STATE //$rand(0,1)$ is the function that random uniformly generate a number from $0$ to $1$.
    \STATE $a = []$;
    \FOR{$i=1$ to $8$}
        \STATE $a.append(rand(0,1))$;
    \ENDFOR    
    \STATE $a.append(1.0)$;
    \STATE $a.sort()$;
    \STATE // Each $c_{k}$ corresponding to a sample distribution.
    \STATE $k=count$;
    \STATE //$f$ is the fractions of response types of individuals, $f[0]=P({y_1}_{x_1},{y_1}_{x_2}|c_k),...,f[8]=P({y_3}_{x_1},{y_3}_{x_2}|c_k).$
    \STATE $f = []$;
    \STATE $f[0] = a[0]$;
    \FOR{$i=1$ to $8$}
        \STATE $f[i] = a[i] - a[i-1]$;
    \ENDFOR
    \STATE // Generate experimental data.
    \STATE $P({y_1}_{x_1}|c_{k})=f[0]+f[1]+f[2]$;
    \STATE $P({y_2}_{x_1}|c_{k})=f[3]+f[4]+f[5]$;
    \STATE $P({y_3}_{x_1}|c_{k})=f[6]+f[7]+f[8]$;
    \STATE $P({y_1}_{x_2}|c_{k})=f[0]+f[3]+f[6]$;
    \STATE $P({y_2}_{x_2}|c_{k})=f[1]+f[4]+f[7]$;
    \STATE $P({y_3}_{x_2}|c_{k})=f[2]+f[5]+f[8]$;
    \STATE // Generate observational data.
    \STATE $P({x_1},{y_1}|c_{k})=rand(0,P({y_1}_{x_1}|c_{k}))$;
    \STATE $P({x_1},{y_2}|c_{k})=rand(0,P({y_2}_{x_1}|c_{k}))$;
    \STATE $P(x_1|c_{k})=rand(P({x_1},{y_1}|c_{k})+P({x_1},{y_2}|c_{k}),\min\{P({x_1},{y_1}|c_{k})+1-P({y_1}_{x_1}|c_{k}),P({x_1},{y_2}|c_{k})+1-P({y_1}_{x_2}|c_{k})\})$;
    \STATE $P({x_1},{y_3}|c_{k})=P(x_1|c_{k})-P({x_1},{y_1}|c_{k})-P({x_1},{y_2}|c_{k})$;
    \STATE $P(x_2|c_{k}) = 1-P(x_1|c_{k})$
    \STATE $P({x_2},{y_1}|c_{k})=rand(0,\min\{P({y_1}_{x_2}|c_{k}),P(x_2|c_{k})\})$;
    \STATE $P({x_2},{y_2}|c_{k})=rand(0,\min\{P({y_2}_{x_2}|c_{k}),P(x_2|c_{k})-P({x_2},{y_1}|c_{k})\})$;
    \STATE $P({x_2},{y_3}|c_{k})=P(x_2|c_{k})-P({x_2},{y_1}|c_{k})-P({x_2},{y_2}|c_{k})$;
    \STATE //Validate the data, the experimental data and observational data should satisfies the following: $P(x,y|c_{k})\le P(y_x|c_{k})\le P(x,y|c_{k})+1-P(x|c_{k})$.
    \STATE $mark = True$
    \FOR{$i=1$ to $3$}
        \FOR{$j=1$ to $2$}
            \IF {$P({y_i}_{x_j}|c_{k})<P(x_j,y_i|c_{k})$ or $P({y_i}_{x_j}|c_{k})>P(x_j,y_i|c_{k})+1-P(x_j|c_{k})$}
                \STATE $mark = False$;
            \ENDIF
        \ENDFOR
    \ENDFOR
    \IF{$mark == False$}
        \STATE $continue$;
    \ENDIF
    \STATE $count = count + 1$;
\ENDWHILE
\end{algorithmic}
\end{algorithm}

\end{document}